\documentclass[sigconf]{acmart}
\AtBeginDocument{%
  \providecommand\BibTeX{{%
    \normalfont B\kern-0.5em{\scshape i\kern-0.25em b}\kern-0.8em\TeX}}}

\usepackage{float}
\usepackage{multirow}
\usepackage{graphicx,subfigure}

\begin{document}

\acmYear{2022}\copyrightyear{2022}
\setcopyright{acmcopyright}
\acmConference[COMPASS '22]{ACM SIGCAS/SIGCHI Conference on Computing and Sustainable Societies}{June 29--July 1, 2022}{Seattle, WA, USA}
\acmBooktitle{ACM SIGCAS/SIGCHI Conference on Computing and Sustainable Societies (COMPASS '22), June 29--July 1, 2022, Seattle, WA, USA}
\acmPrice{15.00}
\acmDOI{10.1145/3530190.3534795}
\acmISBN{978-1-4503-9347-8/22/06}

\title[Use of Metric Learning for the Recognition of Handwritten Digits]{Use of Metric Learning for the Recognition of Handwritten Digits, and its Application to Increase the Outreach of Voice-based Communication Platforms}

\author{Devesh Pant}
\email{devesh.pant@cse.iitd.ac.in}
\affiliation{%
  \institution{IIT Delhi}
    \country{India}
}

\author{Dibyendu Talukder}
\email{dibyendu.t@oniondev.com}
\affiliation{%
  \institution{Gram Vaani}
    \country{India}
    }

\author{Deepak Kumar}
\email{deepak.kumar@oniondev.com}
\affiliation{%
  \institution{Gram Vaani}
  \country{India}
}

\author{Rachit Pandey}
\email{rachit.pandey@oniondev.com}
\affiliation{%
 \institution{Gram Vaani}
    \country{India}
    }

\author{Aaditeshwar Seth}
\email{aseth@cse.iitd.ac.in}
\affiliation{%
  \institution{IIT Delhi and Gram Vaani}
  \country{India}
  }

\author{Chetan Arora}
\email{chetan@cse.iitd.ac.in}
\affiliation{%
  \institution{IIT Delhi}
  \country{India}
}

\renewcommand{\shortauthors}{Pant, et al.}

\begin{abstract}
Initiation, monitoring, and evaluation of development programmes can involve field-based data collection about project activities. This data collection through digital devices may not always be feasible though, for reasons such as unaffordability of smartphones and tablets by field-based cadre, or shortfalls in their training and capacity building. Paper-based data collection has been argued to be more appropriate in several contexts, with automated digitization of the paper forms through OCR (Optical Character Recognition) and OMR (Optical Mark Recognition) techniques. We contribute with providing a large dataset of handwritten digits, and deep learning based models and methods built using this data, that are effective in real-world environments. We demonstrate the deployment of these tools in the context of a maternal and child health and nutrition awareness project, which uses IVR (Interactive Voice Response) systems to provide awareness information to rural women SHG (Self Help Group) members in north India. Paper forms were used to collect phone numbers of the SHG members at scale, which were digitized using the OCR tools developed by us, and used to push almost 4 million phone calls. The data, model, and code have been released in the open-source domain. 
\end{abstract}


\maketitle

\section{Introduction}

In the context of running large development programmes such as SHGs (Self Help Groups) networks, agriculture extension, employment generating activities, and immunization campaigns, among others, it is advisable to maintain digitized records of the programme activities \cite{Dehnavieh2019TheDH, 10.1093/cdn/nzaa053_091_patil}. Digitization allows for ongoing monitoring and learning to understand implementation efficacy, such as the achieved outreach, deviations from planned activities, and other output and outcome indicators. A common strategy is for the programme field teams to use tablets and smartphones to keep a record of their work \cite{10.1093/cdn/nzaa053_091_patil, Saha209}. However, many of these programmes are implemented at the last mile by community-based cadre who may not be very technology savvy, or logistic challenges may arise in organizing training workshops and refresher courses for them, or providing Internet-enabled devices and network packs to the entire field cadre may be expensive \cite{10.1093/cdn/nzaa053_006_Avula, doi:10.1002/itdj.20119_Sundeep, Singhe005041}. As a consequence, the use of traditional paper-based methods has been hard to displace. In fact, in practice, record management for many programmes is done by field workers first on paper, and then retrospectively digitized through data entry into digital devices \cite{MEGHANI2021114291, Gudi164}. 

In our study, we embrace this paradigm of using paper forms as an appropriate method for record keeping by field teams, and demonstrate through a field project of how the benefits of digitization can still be achieved by automatically digitizing paper forms through handwriting recognition using deep learning techniques. 

The relevance of using paper forms has been documented in several contexts. The \emph{Shreddr} system piloted in Mali operated on digital scanned copies of paper forms of surveys about citizen perceptions of governance and accountability of Malian elected officials. It cropped out portions of hand-written text on the forms and passed these ROIs (Regions of Interest) to crowd-sourced platform workers for annotation, while also randomizing the allocation among different workers to preserve privacy \cite{10.1145/2160601.2160605}. This system evolved into a social enterprise, \emph{Captricity}, which now operates in several countries. Another system, \emph{mScan}, was piloted in Mozambique and used OMR (Optimal Mark Recognition) to read data off photographs of custom designed paper forms containing multiple-choice questions \cite{10.1145/2160601.2160604}. The social enterprise, \emph{Health-E-Net}, uses the same concept at scale in Africa, with an innovative method to also mass-produce custom forms by rubber-stamping entire forms on blank paper. 

We draw our motivation from such prior work and contemporary practices, to improve the state of the art in usage contexts where paper-based entry by field workers is easier or more cost effective than to use electronic devices. We go beyond OMR to also recognize hand-written numeric digits. We do not attempt a complete OCR (Optical Character Recognition) solution for alphabets and numerals written in free-text, but only provide a limited solution that allows form designers to build forms with demarcated spaces for single digits, and a dataset and model for recognizing these handwritten digits. Our dataset, consisting of 178,334 digits, was collected through a field project in which paper forms were used to build a phone number database for a large government-run women SHG (Self Help Group) network in the state of Bihar in India. This phone number database was used to send periodic voice messages through an IVR (Interactive Voice Response) system to the SHG members about health and nutrition awareness, including for COVID-19, and other announcements relevant to the SHG programme. We show that this real-world dataset has several nuances because of which models trained on the standard MNIST dataset and its variants do not work well. We also show that commercial OCR APIs by cloud providers like Microsoft Azure and Google Cloud do not perform well on handwritten text. We therefore create our own model trained on our dataset. This model is able to provide a digit-level accuracy of 99.40\% and a phone number (consisting of ten digits) level accuracy of 94.66\% using a metric learning technique with triplet loss \cite{Schroff_2015}. Our trained model is also able to obtain an accuracy of 99.50\% on the MNIST data, validating the generalizability of our model to work on standard datasets in this space. The dataset and model contributed by us can be deployed in similar environments where one-time or ongoing paper-based data collection and record keeping can be done, and digitized in bulk to monitor project activities and milestones. 


Our system is now deployed as part of the voice-based communication platform for SHG members in Bihar, to promote health and nutrition awareness \cite{10.1145/3287098.3287110_behaviour_change, 10.1145/3392561.3394632_managing_participatory}. An SHG consists of 10-12 women, a group of 10-12 SHGs in a village is supported by a community mobilizer, and a group of 30-35 community mobilizers in a village cluster are managed by a cluster head. The cluster heads are informed by our social enterprise partner, Gram Vaani, on how to fill out the paper forms. They in turn inform the community mobilizers and distribute the forms to them. Each community mobilizer then fills a separate form, typically one for each SHG, with the phone numbers of the SHG members. These filled forms are then collected by the cluster heads and eventually passed on to the Gram Vaani team, who scans them in bulk, and runs the scanned forms through the OCR system. 

\begin{figure*}[t]
	\begin{center}
		\includegraphics[width=\textwidth,
		height=\textheight,keepaspectratio]{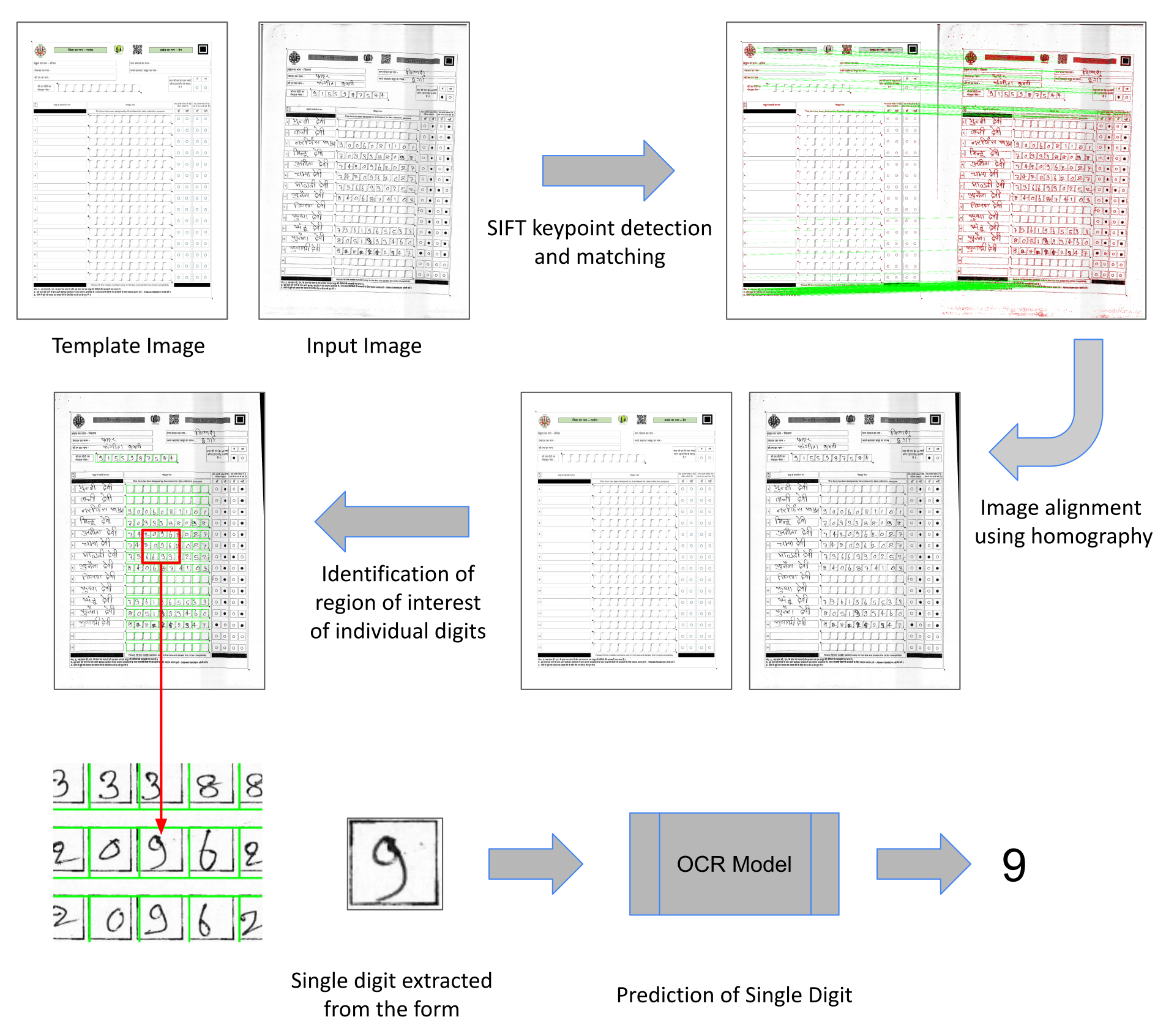}
		\caption{OCR pipeline}
		\label{fig:fig1}
	\end{center}
\end{figure*}

A high-level pipeline of our OCR system is shown in Figure \ref{fig:fig1}. A homography \cite{HZbook} step first normalizes the form to correct for any distortions created in the scanning process, then individual digits are cropped according to the form template, and finally each digit is classified using a deep learning model. Digits predicted with a low confidence score can also be flagged and passed on for manual correction by data entry operators. The phone numbers finally collected through this process, are then used to push awareness based voice messages to the SHG members through Gram Vaani's IVR system. 

We also show that this process to collect phone numbers and send awareness-based voice messages is a viable means to address several aspects of the digital gender divide. It is well known that access and ownership of phones by women is much lower than that of men \cite{1343822_a_tough_call}. Further, even if rural women do have access to mobile phones, they may have limited skills to use them, often restricted to being able to receive calls but not to make calls by dialing phone numbers or accessing the address-book \cite{10.1145/3287098.3287110_behaviour_change}. Their phones are also often \emph{hand-me-downs} from male members of the household, and may not be in good shape, with worn out keypads, jammed keys, broken screens, etc \cite{10.1145/3460112.3471963}. According to the original design of the voice-based communication awareness project by Gram Vaani, several such challenges were addressed by persuading women to negotiate access to phones in their household, and by imparting training to them including on basic mobile literacy and on how to use the IVR system, so that they could access the information on-demand by calling the unique toll-free phone number for the IVR system. It was found that the system usage went up with repeated trainings and the strong solidarity within the SHG groups to collectively use the platform to learn and share their experiences \cite{10.1145/3287098.3287110_behaviour_change}. However, as the system was scaled to more districts and the training process was transitioned to the government department running the SHG network, such an intensive training became harder. In the absence of such training, the system usage remained low in new geographies where it was expanded, and a method was therefore needed to collect the phone numbers of the SHG women at scale so that instead of having them access the IVR system on their own (pull-based), messages could be periodically pushed towards them (push-based). In this study, we also provide a comparison between the push-based and pull-based models through usage statistics of the IVR, to compare the programme efficacy between the two models. 

Our contributions include releasing a large labeled dataset of handwritten digits cropped from paper forms, high-accuracy models for the recognition of these digits that perform better than alternate state-of-the-art methods, and code and library provided in the open-source for social development practitioners to build and digitize their own paper-based forms. The code and data can be obtained from \footnote{\url{https://github.com/Smartforms2022/Smartforms}}. 

\section{Related Work}
\begin{figure}[!hbt]
\begin{center}
  \includegraphics[width=1\linewidth]{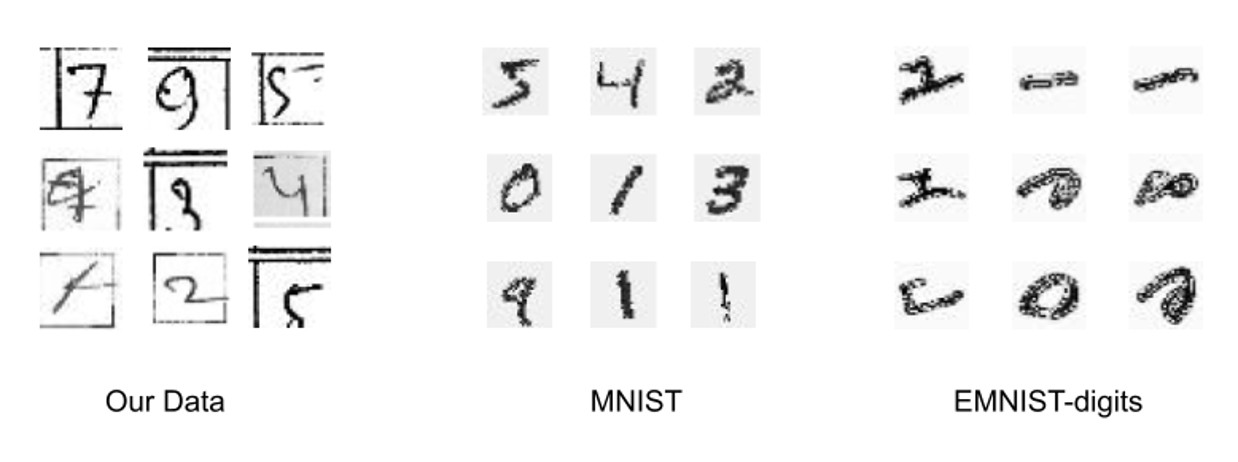}
  \caption{The digits in the MNIST and EMNIST datasets are centred, size corrected, and with no noise; however, our dataset has border noise and several digits are partially cropped due to homography inaccuracy.}
  \label{fig:mnist_data}
\end{center}
\end{figure}

Handwritten digit recognition is a popular task in computer vision, and CNN (Convolutional Neural Network) based methods are known to deliver a good performance. LeCun \emph{et al}. \cite{726791} proposed the LeNet-5 CNN architecture which performed well on the MNIST dataset, with an accuracy of 99.05\% \cite{lecun-mnisthandwrittendigit-2010}. More recent work by Yoon \emph{et al}. \cite{s20123344} achieved an accuracy of 99.87\% with a close investigation on the role of various hyper-parameters such as the numbers of layers, stride size, kernel size, padding and dilution, etc. We however find that real-world datasets have certain peculiarities such as noise introduced due to imprecise cropping, data imbalance, and cases of high intra-class variance as well as high inter-class similarity (shown in Figure \ref{fig:mnist_data}), for which models trained using metric learning based methods are able to get a better representation of the data and hence a better classification performance\cite{sym11091066}. Instead of direct classification of the output classes from the CNN architecture, metric learning aims to learn a representational embedding such that similar images are close to one-other and dissimilar images are far apart in the embedding space \cite{DBLP:journals/corr/abs-2003-08505}. Embedding loss functions such as Contrastive Loss \cite{1467314} and Triplet Loss \cite{Schroff_2015} are examples of this method. Triplet loss-based metric learning has demonstrated superior performance than contrastive loss in a variety of recognition tasks. Schroff \emph{et al.}, for example, improved face recognition performance by employing triplet loss-based metric learning \cite{Schroff_2015}. Thapar \emph{et al.} used triplet loss to recognise camera-wearing subjects in egocentric videos \cite{Thapar2020IsSO}. Zeng et al. \emph{et al.} incorporated triplet loss in CNN architectures for handwritten Chinese character recognition \cite{zeng_8270001}. We similarly show that the triplet loss function provides us with good performance on our real-world dataset, and performs better than direct classification models. 

Note that our approach for hand-written digit recognition is distinct from another popular domain where OCR is used: scene-text recognition \cite{https://doi.org/10.48550/arxiv.2005.03492}. Scene-text recognition is applied on images of signboards, billboards, vehicle number plates, etc., and solves a different set of problems such as to correct for image distortions, complex backgrounds, and varying shape and sized fonts. In the document OCR problem domain that we address, we perform text detection and localization using homography matching against a pre-defined template, and then focus just on the hand-written digit recognition using CNN-based methods. 

With our dataset and model for handwritten digit recognition, we are able to contribute to the broad goal of digitization of paper forms. As mentioned earlier, we are able to go beyond straightforward OMR based tasks which restrict the question space to only multiple-choice based options \cite{10.1145/2160601.2160604}, and cropping tasks which commission the annotation of ROIs (Regions of Interest) to crowd-sourced workers \cite{10.1145/2160601.2160605_shreddr}. We are in the process of releasing an open-source library, along with our dataset and model, so that form designers can build custom forms with OMR and handwritten digit OCR fields. 

Having digital interfaces resemble their paper-based counterparts has also been an area of investigation. The CAM system annotated paper forms with unique \emph{QR code}-like IDs for different fields, which when photographed by field workers assisted them through voice prompts to type in the data correctly \cite{inproceedings@parikh_tapan}. Digital slates with handwriting recognition were piloted for record-keeping by allowing field workers to provide information as they would on an equivalent paper form \cite{managing_microfinance_with_paper}. The Partograph is a standardized tool for birth attendants to monitor cervical dilation and other indicators during the birthing process and use it for decision support; the Partopen system printed the Partograph on dotted paper so that a digital pen could be used to record the data digitally and provide real-time decision support through standardized health protocols \cite{inproceedings_underwood}. While these tools echo the relevance of paper-based interfaces, they are expected to be more expensive involving the purchase of digital devices and consequent training of the field workers to use them. We therefore believe that there continues to be a space of tasks where traditional paper forms are more appropriate than switching to digital devices, and handwriting recognition solutions such as the one developed by us are relevant in such settings.

\begin{figure}[!hbt]
\begin{center}
  \includegraphics[width=1\linewidth]{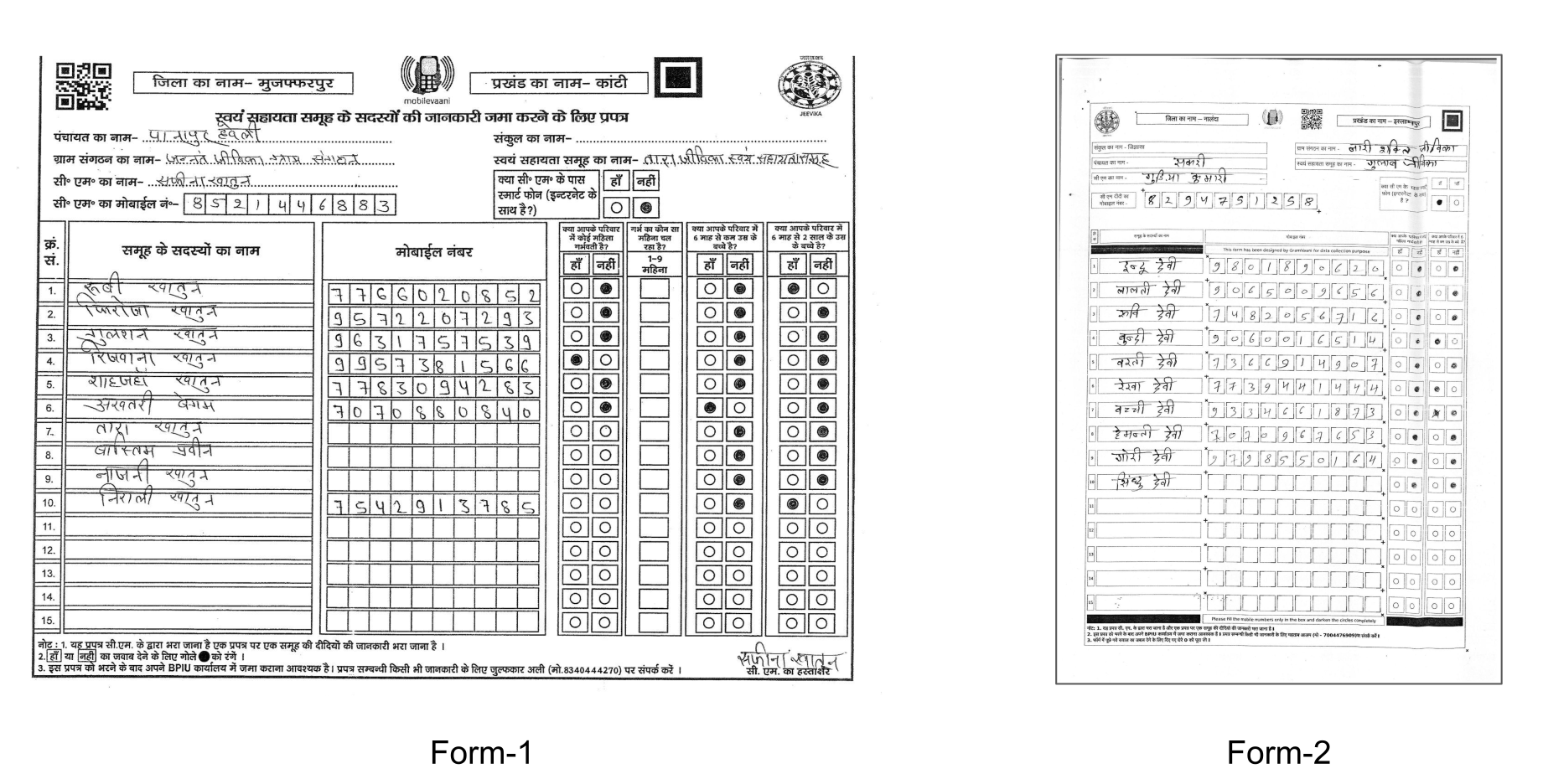}
  \caption{Form variants used in the field}
  \label{fig:fig2}
\end{center}
\end{figure}

Figure \ref{fig:fig2} shows two versions of the paper forms used by us in the field. Separate forms are printed for each sub-district, with a unique QR-code for each sub-district, so that segregation of the filled forms during the bulk scanning process need not be maintained. Each form contains 16 rows to collect the phone numbers for up to 16 members in an SHG. The names of the members are ignored, only the phone number digits are extracted for recognition. Additional OMR columns are also used to collect data such as whether the SHG member has a pregnant woman in the family, or a child that is less than 6 months old, or a child that is less than two years old. This data was intended to be used to push customized voice messages to the SHG families, but did not end up getting used because the content developed for the project was story-oriented and stretched across multiple audio pieces, rather than each piece being self-contained in itself. Therefore it was eventually decided to push all voice messages to everybody. The forms also contained a separate field for the phone number of the community mobilizer, so that it could be used to relate the IVR system usage by SHG members with their corresponding community mobilizer. Usage statistics of the IVR system could then be used to build performance metrics for the community mobilizers based on the system usage observed from among the SHG members in their groups. This could even be aggregated to the village cluster level, and further to the sub-district level, to monitor system use and provide targeted guidance to the cluster heads or community mobilizers if needed. 

In the context of the OCR system, the two versions of the forms primarily differed in terms of the outline of the cells for each phone number digit. Form-1 had darker boundaries than Form-2, which we used to evaluate whether noise introduced by the boundaries made the OCR task harder on Form-1. Form-2 also had special marks such as `$+$' and `$x$' on the cell corners, which we introduced with the presumption that it may help us improve the homography through local transformations in the cell neighbourhood in case it was needed. The forms were scanned and normalized to produce cells of size 32x32 pixels. Cells obtained from approximately 1000 forms of each Form-1 and Form-2 were annotated by a contracted data entry firm, with each digit being annotated independently by two operators to ensure consistent labeling. Overall, 989 Form-1s were annotated to obtain 74,595 labels for non-blank cells, and 990 Form-2s were annotated to obtain 103,739 labels for non-blank cells. 

\section{Methodology and Results}

\subsection{Homography}

To normalize the scanned form images against a template image (of a blank form), we first apply the SIFT (Scale-Invariant Feature Transform) algorithm \cite{ctx1645550150006836} to find key-points on both the images, match them together, and then compute the homography matrix over the matched key-points. A perspective transformation is then applied to align the scanned image with the template image, and the digit ROIs are extracted.

\begin{figure}[t]
\centering     
\subfigure[]{\label{fig:4a}\includegraphics[width=70mm]{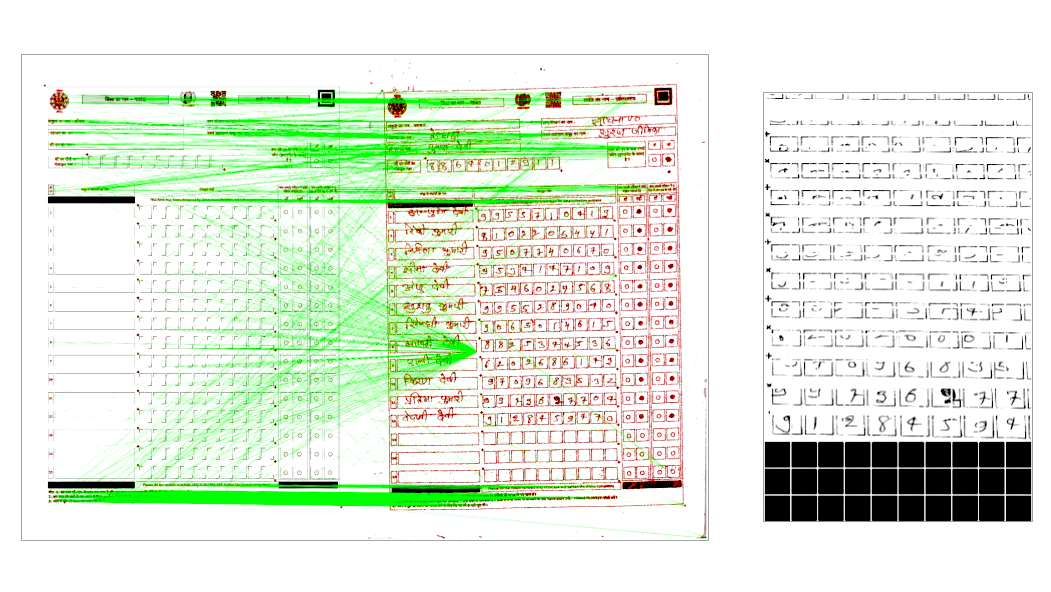}}\hspace{5mm}
\subfigure[]{\label{fig:4b}\includegraphics[width=70mm]{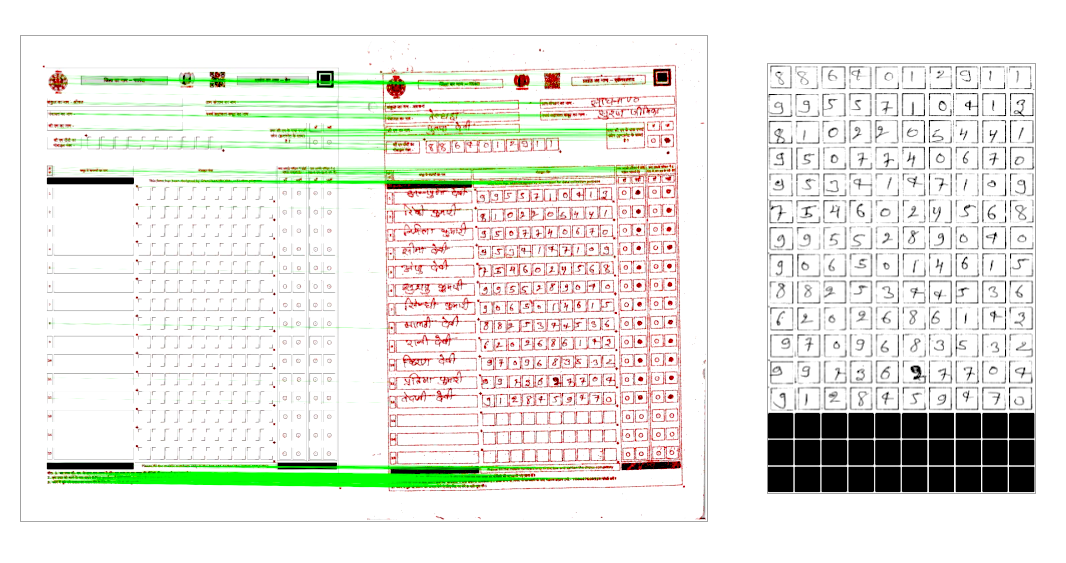}}
\caption{\textbf{(a)} Several incorrect key-points are matched, \textbf{(b)} Several incorrect mismatches are eliminated by assuming (approximate) vertical alignment in the scanning process}
\label{fig:fig4}
\end{figure}

Figure \ref{fig:fig4}(a) shows that an initial attempt to match all key-points led to many incorrect matchings. To quantify the quality of the homography, we manually created a small dataset of 400 bounding boxes of digit cells randomly selected across 25 randomly selected forms, and computed the mean IoU (Intersection over Union) scores for these cells. The IoU between bounding boxes A and B, from the manually marked and automatically inferred bounding boxes, is computed as: 
\[
    IoU = \dfrac{\text{Area of Intersection}(A \cap B)}{\text{Total Area}(A \cup B)}
\]
The mean IoU scores obtained were 88.21\% and 62.39\% for Form-1 and Form-2 respectively, which when applied to the baseline OCR model (as explained in the following section) led to a single-digit prediction accuracy of 97.74\%. To improve the homography, we imposed a threshold to only match key-points if their coordinates $(x, y)$ on the scanned image and coordinates $(x', y')$ on the template image did not differ vertically beyond a particular threshold, i.e. $abs(y'- y) <= threshold$, with the assumption that most forms would be scanned with their right-side up. Figure \ref{fig:fig4}(b) shows that this step helped improve the homography by eliminating several spurious mismatches, when using a threshold of 150 pixels (on scanned images with a resolution of 4134x5846 pixels). The threshold can vary based on the image resolution. We evaluated several different threshold settings and chose the one that produces the best mean IoU score. This selective keypoint matching method helped improve the IoU score of Form-2 to 77.50\%, and led to an overall single digit prediction accuracy of 99.12\%. We therefore proceed with this choice of threshold for the subsequent OCR step.

\subsection{Blank classification of digit cells}

Before the OCR step, we need to identify which digit cells contain handwritten digits and which ones are blank. We create a binary classifier for this purpose, using an architecture similar to the LeNet-5 architecture for digit classification. The dataset consists of approximately 60,000 equally balanced blank and digit cell images, sampled evenly from both types of forms. We divide it into a train, validation, and test split of 60:20:20. The model is trained for 30 epochs and achieves a test accuracy of 99.99\% on Form-1 and 99.98\% on Form-2. The false positive rate (predicting a non-blank digit cell as blank) on both the forms is 0, and false negative rates (predicting a blank as a non-blank digit cell) are 0.0001 and 0.0002 on Form-1 and Form-2 respectively. These false negative rates are eliminated once predictions for other digits of the same phone number are also taken into account, to allow a non-blank prediction only if at least one more cell in the same row is also predicted as non-blank. This improves the accuracy to 100\% and non-blank cells thus identified are sent to the subsequent OCR step for digitization. 


\subsection{OCR: Direct classification model}

As a baseline, we create a CNN model based on the LeNet-5 architecture \cite{726791} and train the model using a softmax classification loss function. The architecture consists of five convolution layers with $3\times3$ kernels for feature extraction, each followed by a max-pooling layer of size 2 and a batch normalization layer. Dense layers are added after the convolution layers and finally a softmax function gives the final class probabilities. We use ReLU as the activation function and also add a dropout of factor 0.5 to the dense layers. 

The dataset is divided into train, validation, and test splits of 60:20:20. Transformations such as random cropping are applied to the input images. We use a batch size of 256, and Adam \cite{kingma2017adam} as the optimizer. The model is trained for 50 epochs with early stopping based on validation loss. We start with the learning rate of 0.001 which is lowered as the training progresses. 

\begin{figure}[t]
\begin{center}
   \includegraphics[scale=0.40]{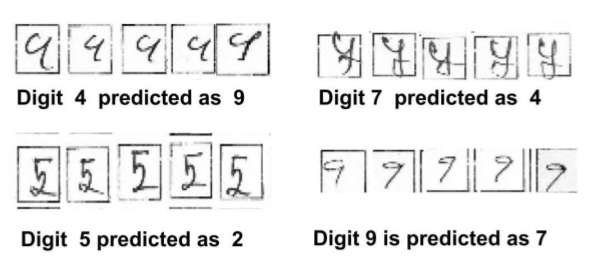}
   \caption{Example of digits misclassified by the direct classification model}
   \label{fig:fig5}
\end{center}

\end{figure}

\begin{figure}[!hbt]
\begin{center}
   \includegraphics[scale=0.20]{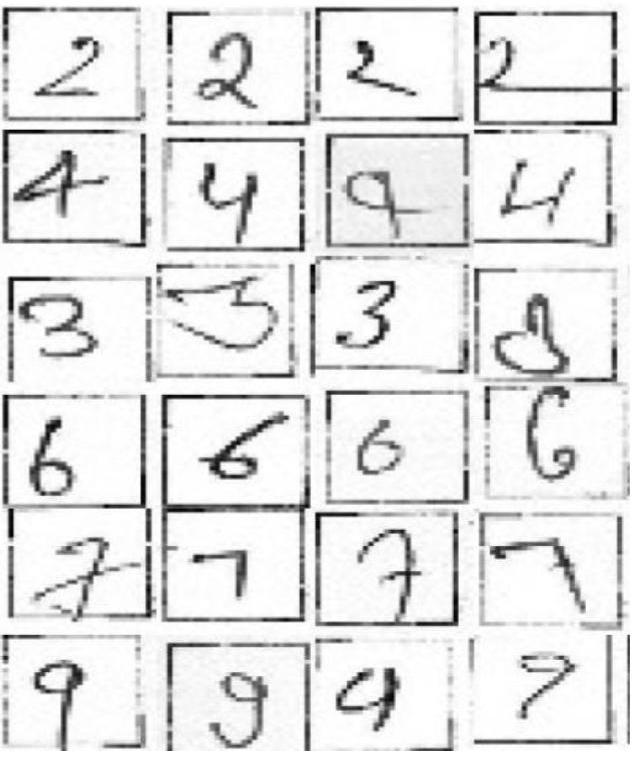}
   \caption{Different handwriting styles for same digits}
   \label{fig:fig6}
\end{center}

\end{figure}

Table \ref{tab:table-1} shows the digit-level and phone number level (all ten digits are correct) accuracy for both the form versions. An error analysis shown in Figures \ref{fig:fig5} and \ref{fig:fig6} reveals some common sources of mis-classification. On the one hand, some digits belonging to different classes seem quite similar to one another. On the other hand, some digits seem to have quite dissimilar writing styles among different people. This leads us to try metric learning based approaches, as explained next.

We also try pre-trained CNN models such as VGG-16 \cite{simonyan2015deep} but do not achieve any significant improvement in the test accuracy, as shown in Table \ref{tab:table-1}. We therefore move towards improving the LeNet-5 direct classification method to a metric learning based method.

\subsection{OCR: Robust embedding using Metric Learning and Triplet loss}

Metric learning tries to learn a representation (embedding) for each image such that similar images are close to one-other and dissimilar images are far apart in the embedding space \cite{DBLP:journals/corr/abs-2003-08505}. Metric learning seems more suitable in our scenario to address the challenges depicted in Figures \ref{fig:fig4} and \ref{fig:fig5}. Instead of training a classifier with cross entropy loss, we first learn an embedding vector using a triplet loss function. Once the embeddings have been learned, a softmax based classification layer is added to give the confidence score prediction for each of the 10 digit classes. The digit with the highest score is used as the prediction class. 

The triplet loss function for learning the representational embeddings selects three kinds of samples from the training set: \emph{anchor} (a), \emph{positive} (p), and \emph{negative} (n). Anchor and positive samples are chosen such that they belong to the same class, whereas a negative sample belongs to a different class than the anchor and positive. The triplet loss function minimizes the distance between the anchor and positive samples to achieve intra-class compactness, as well as, maximizes the distance between the anchor and negative samples, to achieve greater inter-class discrimination. The triplet loss function can be defined as:

\[ 
\text{Loss}(a,p,n) = \max{(\| a-p\|^{2} - \|a-n\|^{2} + \alpha, 0)}, \text{ where } \alpha = \text{margin}.
\]

Schroff \emph{et al.} suggest that the triplet loss function should be computed on a subset of samples instead of all the samples. \cite{Schroff_2015}. They distinguish between three kinds of samples: \textbf{easy triplets} for which the positive samples are closer to the anchor within the margin, and thus the loss is always zero 
($\| a-p\|^{2}  + \alpha <=  \|a-n\|^{2}$), \textbf{hard triplets} for which the negative samples are closer to the anchor than the positive samples
($\|a-n\|^{2} < \| a-p\|^{2}$), and \textbf{semi-hard triplets} for which within the margin the negative samples are closer to the anchor than the positive samples, to make the loss positive ($\|a-n\|^{2} <= \| a-p\|^{2}  + \alpha$). Semi-hard triplets gave better results for us: For every batch of inputs, we obtain the embedding vectors, select only the valid triplets according to the semi-hard condition, and compute the loss on these triplets. 

This representational learning architecture has some additional minor differences from the direct classification architecture. We remove all the batch normalization layers to reduce overfitting. A dense layer of 128 output units is added after the five convolution layers and a dropout factor of 0.5 is incorporated in this layer. The input images are mapped to a 128 dimensional Euclidean representation, over which the triplet loss is calculated. Several values for the margin $\alpha$ were tested, with $\alpha = 0.1$ giving the best results. The model is trained for a maximum of 50 epochs with early stopping on validation loss which stabilizes after 40 epochs. 


\begin{table*}[!hbt]
    \centering
  \begin{tabular}{|l|l|l|l|l|l|l|c}
  \hline
    \multirow{2}{*}{Method} &
      \multicolumn{2}{c|}{Form-1 (\%)} &
      \multicolumn{2}{c|}{Form-2 (\%)} &
      \multicolumn{1}{c|}{MNIST (\%)} &
      \multicolumn{1}{c|}{EMNIST (\%)} \\
       \cline{2-7}
       & {Digit} & {Phone} & {Digit} & {Phone} & {Digit} & {Digit} \\
       \hline
       LeNet-5 based direct classification model  & 99.14 & 93.50 & 99.09 & 92.02 &     & \\
       VGG-16 based direct classification model & 99.18 & 93.62 & 99.11& 92.04& & \\
       Triplet loss based model  & 99.35 & 94.23 & 99.12 & 92.22 & 99.50 & 99.62\\
       Triplet model trained on MNIST data & 65.24 &     & 65.35 &      &     & \\
       Triplet model trained on EMNIST data & 12.14 &    & 14.80 &      &     &\\
       Azure OCR, applied on the entire grid    & 23.87 & 6.12  & 26.00   & 8.23   & & \\
       Azure OCR, applied on individual digits  & 36.34 & 10.05 & 38.93 & 10.67  &  &\\
       Google Vision, applied on the entire grid & 20.02 & 3.80 & 39.60 & 11.80 & &\\
       Google Vision, applied on individual digits &  34.60 & 9.67 & 37.74 & 10.24& &\\
       Paddle OCR, applied on the entire grid &  &  & 33.56 & & &\\
       
     \hline
  \end{tabular}
  \newline
  \caption{\label{tab:table-1}
Performance comparison of different models on various datasets}
\end{table*}

\begin{figure*}[!hbt]
\centering     
\subfigure[Softmax ]{\label{fig:7a}\includegraphics[width=60mm]{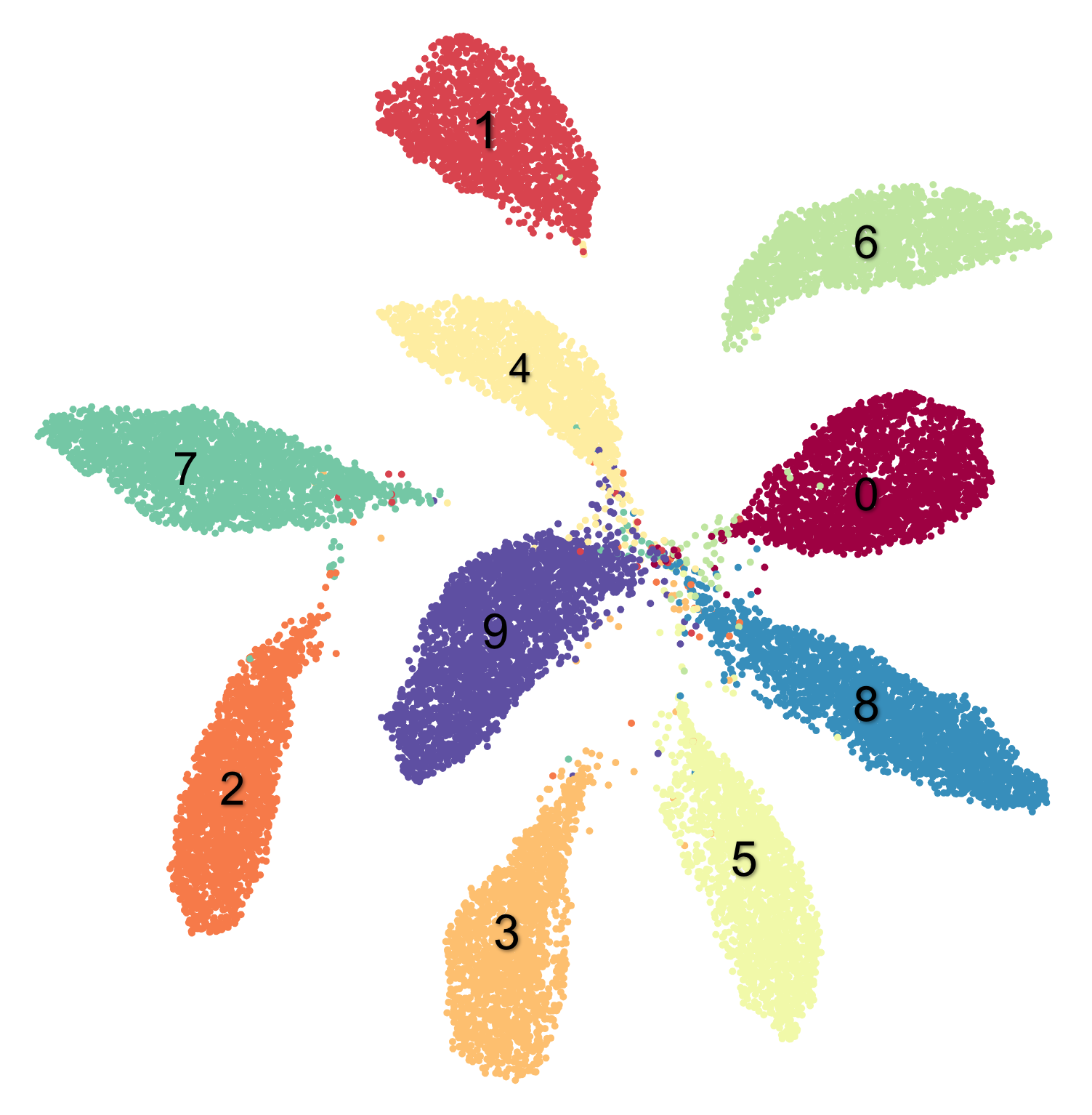}}\hspace{10mm}
\subfigure[Triplet]{\label{fig:7b}\includegraphics[width=60mm]{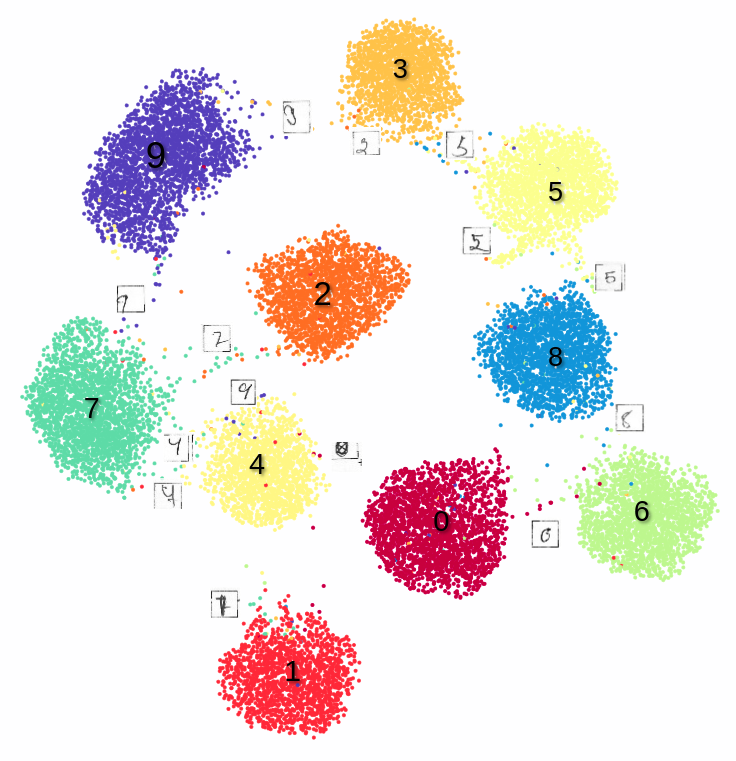}}
\caption{2D visualization of the embeddings}
\label{fig:fig7}
\end{figure*}

Table \ref{tab:table-1} reports the performance of different models on digit-level and phone-level accuracy. A phone number prediction is considered correct when all its ten digits are correctly recognized. We can see that the triplet loss based model performs better on both metrics than the direct classification model. Most of the erroneous cases shown in Figures \ref{fig:fig4} and \ref{fig:fig5} are correctly classified with this model. UMAP plots of the embeddings obtained through both the models are shown in Figure \ref{fig:fig7}: the triplet loss model is able to achieve a better class separation. We also found that 92\% of incorrectly predicted phone numbers in Form-1 and 87\% of incorrectly predicted phone numbers in Form-2 just had one mis-predicted digit. Errors in the triplet loss model therefore seem to be uncorrelated.

We also run this model on the MNIST dataset \cite{lecun-mnisthandwrittendigit-2010}, and its extended version of the EMNIST dataset \cite{cohen2017emnist}. We find that the model is able to generalize to these datasets with performance comparable to the state of the art models demonstrated on these datasets (last two columns in Table \ref{tab:table-1}). 

To additionally validate the relevance of our dataset, we train the same triplet loss model on the MNIST and EMINST datasets, and evaluate its performance on our real-world dataset. The results in Table \ref{tab:table-1} show that the models trained on these cleaner datasets are not able to perform as well on real-world data.

\subsection{Benchmarking against commercial OCR systems}

We also evaluate the OCR APIs provided by the Microsoft Azure and Google Cloud platforms. We first pass the entire grid of 16 phone numbers to these platforms to see if their text-spotting models are able to detect all the digits. We find that the Azure OCR is unable to detect all the digits while the Google Cloud OCR is able to detect most of the digits but gets the boundaries incorrect, as a result of which the prediction accuracy is low. 

For a direct comparison on digit-level accuracy, we use our homography routines to extract individual digits and send a random sample of 1600 digits to the Azure and Google Cloud OCR APIs. The results improve for both the models, but the accuracy remains much lower in comparison with what we are able to achieve (Table \ref{tab:table-1}). 

We similarly evaluate a pre-trained text spotting and recognition model provided by the PaddleOCR Library \cite{https://doi.org/10.48550/arxiv.2009.09941}.  The model was unable to detect digits in Form-1 where the neighbouring digit cells are joined with one another, but detected 60\% of digits in Form-2. The accuracy of single digit recognition was only 33\% and this led to zero accuracy at the phone number level. This demonstrates that our method of building a pipeline with homography followed by digit recognition performs much better than text-spotting methods.

\subsection{Segregation for manual annotation}

We next attempt to further improve the accuracy by targeted manual correction of the model predictions. Using the confidence scores for digit level predictions, we experiment with different thresholds to divide the input data into a low confidence group (confidence \(<\) threshold) and a high confidence group (confidence \(\ge\) threshold). The low confidence group forms less than 3\% of the overall data for different threshold values. We consider the scenario that all digits in this group are put up for manual annotation to data entry operators. Table \ref{tab:table-2} shows the performance, for the low confidence group in terms of the original accuracy which is improved to 100\% through the manual corrections, for the high confidence group by taking the predicted values directly as such, and the overall accuracy. This targeted correction method improves the phone number level accuracy significantly, to more than 98\% for Form-1 and 95\% for Form-2. Table \ref{tab:table-2} also shows the additional cost estimate per 10,000 forms for this manual annotation, and the time taken for one data entry operators to resolve these flagged digits. These cost and time estimates are based on payments made and discussions between Gram Vaani and the contracted data entry firm. Form-1s used in Muzaffarpur had on average 9.39 ($\sigma = 3.11$) phone numbers per form, whereas Form-2s used in Nalanda had on average 9.56 ($\sigma = 2.76$) phone numbers per form. The cost of manual annotation of a form is INR 8.5 (USD 0.11) with one data entry operator able to annotate 75 forms in a day on average. For 10,000 forms, even with the choice of a low confidence threshold the cost ranges at around INR 20,000 (USD 250) and the corrections can be done by a single data entry operator in about four days. 

\begin{table*}[!hbt]
\centering
\begin{tabular}{|c|c|c|c|c|c|c|c|}
 \hline
        \multicolumn{3}{|c|}{\vtop{\hbox{\strut\textbf{Low Confidence}}
        \hbox{\textbf{\strut Group}}}}&
        
        \vtop{\hbox{\strut\textbf{High}}\hbox{\strut\textbf{Confidence}}
        \hbox{\textbf{\strut Group:}} \hbox{\textbf{\strut Accuracy (\%)}}}&
        \vtop{\hbox{\strut\textbf{Overall}}\hbox{\strut\textbf{digit-level}}\hbox{\strut\textbf{accuracy (\%)}}}&
        \vtop{\hbox{\strut\textbf{Overall}}\hbox{\strut\textbf{phone number}}\hbox{\strut\textbf{accuracy(\%)}}}&
        \vtop{\hbox{\strut\textbf{Cost per}}\hbox{\strut\textbf{10,000 }}\hbox{\strut\textbf{forms}}\hbox{\strut\textbf{ (INR)}}}&
        \vtop{\hbox{\strut\textbf{Time taken}}\hbox{\strut\textbf{per 10,000}}\hbox{\strut\textbf{forms (days)}}\hbox{\strut\textbf{for correction}}}\\ 
        \cline{1-3}
        {\textbf{Threshold}}&
        \vtop{\hbox{\strut\textbf{\space\space\%}}\hbox{\strut\textbf{Data}}\hbox{\strut\textbf{ }}}&
        \vtop{\hbox{\strut\textbf{Initial}}\hbox{\strut\textbf{accuracy}} \hbox{\strut\textbf{(\%)}}}&&&&&
        \\
        \hline
        
        \multicolumn{8}{|c|}{\textbf{Form-1}}\\ 
        \cline{1-8}
    
        0.6& 0.86& 60.93& 99.48& 99.49& 95.93 & 6,450& 1.1\\
        0.7& 1.51& 63.55& 99.69& 99.70& 96.77 & 11,325& 2.0\\
        0.8& 2.19& 71.16& 99.78& 99.78& 97.45 & 16,425& 2.9\\
        0.9& 3.45& 78.36& 99.89& 99.90& 98.47 & 25,875& 4.6\\

    \hline
    \hline
    \multicolumn{8}{|c|}{\textbf{Form-2}}\\ 
    \cline{1-8}
        \hline
        0.6& 0.47& 47.12& 99.32& 99.34& 93.33 & 3,525& 0.6\\
        0.7& 0.89& 54.90& 99.46& 99.46& 94.03 & 6,675& 1.2\\
        0.8& 1.44& 60.74& 99.62& 99.63& 95.12 & 10,800& 1.9\\
        0.9& 2.37& 69.85& 99.77& 99.79& 95.52 & 17,775& 3.2\\
        
    \hline

\end{tabular}
\newline
\caption{\label{tab:table-2}
Accuracy improvement and cost implication with manual correction of low-confidence predictions}
\end{table*}

\subsection{Further observations on form design}
As mentioned earlier, we designed Form-2 with some intuitive modifications made on Form-1 to improve homography, but these modifications in fact degraded the performance. In Form-1, we used a dark outline to segregate the grid-cell for each digit, and adjacent cells were joined with one another. In Form-2, we used a lighter outline and also introduced a gap between adjacent cells. We found however that these changes did not improve the performance and noise due to cropped borders continued to affect the performance in both the forms. 

The second change we had introduced was to use marks such as `$+$' and `$x$' on the cell corners in Form-2 with the expectation that it would enhance the SIFT keypoints detection and improve homography. However, we found that these marks in one row were incorrectly matched with similar marks in other rows, resulting in an increase in homography inaccuracy (also shown in Figure-\ref{fig:4b}). Perhaps including unique markers on different rows may improve the performance, and we plan to experiment with this in the future.

\section{Project experience}

During 2019-2020 before the COVID-19 pandemic hit and impacted field operations, Gram Vaani rolled out Form-1 in parts of the district of Muzaffarpur in Bihar, and Form-2 in parts of the district of Nalanda. In total, 2749 SHG community mobilizers participated in the exercise, leading to the collection of 28,915 forms. As described earlier, approximately 1000 each of Form-1s and Form-2s were annotated in duplicate by a contracted data entry firm, and used to train and test the OCR models evaluated as described in the previous section. The voice-based communication project is now slated for scale-up to four additional districts, and eventually to the entire state of Bihar, with the process of building phone number databases through OCR-ed paper forms as an integral part of the project. 

We next describe in brief the project experience with using the phone number database to push awareness information to SHG members in the first two districts of Muzaffarpur and Nalanda. Our goal is to understand whether this method of collecting phone numbers and pushing voice messages to people is a viable mechanism to create awareness. This is because a significant concern with the push-based approach viz-a-vis. a pull-based approach is that push-based calls may get missed in case the recipients are busy when they receive the calls. They may not pick up, or pick up but quickly close the call. Pull-based approaches where people call the toll-free phone number to listen to content on-demand, are therefore considered better, although as discussed earlier, creating awareness and building the capacity of people to call on-demand is much harder than the push-based model. To compare the two models, we define two groups of users, with each user identified by a unique caller-id: 

\begin{itemize}
    \item \emph{Call-in users}: These are users whose first interaction with the IVR platform was initiated by them, i.e. the Gram Vaani field team or trained community mobilizers told them about the IVR platform through in-person trainings and workshops, after which they called the toll-free phone number of the IVR platform. 
    \item \emph{OCR users}: These are users whose first interaction with the IVR platform was initiated by the Gram Vaani IVR system, i.e. their phone numbers were obtained through paper forms, and calls were then pushed to them. 
\end{itemize}

We did this segregation based on the first IVR interaction with a user because both pathways of creating community outreach, i.e. through trainings of the SHG members and through collection of their phone numbers on paper forms, were supposed to happen in all locations but their sequencing could not be ensured because it was driven by staff availability and programme priorities in different locations. Hence we could not assume, for instance, that a caller-id collected via the paper form was not already aware beforehand about the IVR platform from having attended a training in their community in the past. This method of segregation does have some limitations though. The SHG members may report a different phone number in the paper forms, perhaps with a conservative mindset to use their husband's phone number for any formal documentation, and this phone number would then appear as a new user in the OCR group. Multiple household members may also access the IVR platform from the same phone number, and we will not be able to tell from the caller-id alone that these accesses were by different people. Notwithstanding these limitations, we feel that it is reasonable to assume that such cases would be more of exceptions than the norm, and that segregating by the first IVR-caller-id interaction into \emph{call-in users} and \emph{OCR users} will be a satisfactory indicator of how a user entered the system. 

Table \ref{tab:table-3} shows the number of users in each group. The difference between the relative group sizes in Nalanda and Muzaffarpur is due to the different sequencing of field activities in the two districts. Nalanda saw an early pilot of the IVR system during 2016-2018 when the outreach method was exclusively through offline trainings and workshops; paper forms to collect phone numbers for more exhaustive coverage were introduced much later and targeted in sub-districts where offline trainings had not happened. Muzaffarpur was added as a new district in 2019, and was instead jumpstarted through phone number databases collected through paper forms, while offline trainings were simultaneously and gradually scaled up in some sub-districts. 

We then compare the two groups of \emph{call-in users} and \emph{OCR users} on the following indicators: 
\begin{itemize}
    \item \emph{Fraction of push-based calls answered}: This will help understand if the \emph{call-in users} who have a better awareness beforehand about the platform, tend to pick-up push-based calls more often than \emph{OCR users}, who may not have any prior context about the platform when they receive these calls. We count a push-based call as having been answered only if it was picked-up and heard for at least 30 seconds. 
    \item \emph{Push-based mean call duration}: With a similar intention as above, we study if the two groups exhibit any differences in terms of the duration for which they listen to content played over the push-based calls.
    \item \emph{Pull-based mean call duration}: Similarly, we compare the listening duration for pull-based calls made by the two groups to the toll-free IVR system. Note that in all push-based calls, the toll-free IVR number is announced to the recipients so that over time they can learn about this phone number to call anytime on-demand. In Table \ref{tab:table-3} we also mention the percentage of such \emph{OCR users} who eventually transitioned to pull-based on-demand calling, either by coming to know about it from the information that was played to them, or through trainings that they may have attended after already having been included in the database through the OCR-ed forms. 
\end{itemize}

Table \ref{tab:table-3} shows the comparison between the two groups, computed on IVR usage statistics between the period of April 2020 to September 2021. This period included an intense messaging phase between June to August 2020 when the COVID-19 lockdown was imposed in India and several awareness messages related to safe behavior, social protection entitlements, and building back lost livelihoods, were sent to the SHG members \cite{10.1093/heapro/daab050}. We focus on a comparison between the two groups of users based on the three indicators. 

We find that \emph{OCR users} tend to answer push-based calls more often than \emph{call-in users}. This is expected, either due to a preference of call-in users to call on-demand if they can, or these users may have beforehand heard the same content over pull-based calls since content is published there as soon as it is ready whereas it is pushed through outbound calls often a few days later according to a weekly schedule. Similarly, the duration of listening to push-based calls is slightly higher for \emph{OCR users} than for \emph{call-in users}. The length of audio content pieces pushed over these calls are in the range of 1-3 minutes, hence a mean listening duration of approximately 2.5 minutes per call is non-trivial and shows that the overall approach of paper-based phone number collection followed by pushing awareness messages seems reasonable. The listening duration over pull-based calls is much larger than push-based calls though, ranging from 6.5 minutes in Muzaffarpur to more than 8 minutes in Nalanda, and shows that the push-based approach should indeed be seen as a stop-gap and not a substitute for offline training. 

This is further reinforced when we observe that very few \emph{OCR users} transitioned to placing calls on-demand themselves. In Nalanda, this percentage is 4.2\% whereas in Muzaffarpur the percentage is only 1.6\%. This indicates the importance of offline trainings and workshops to bridge the digital gender divide more effectively, however in its absence due to various logistic and cost issues as explained, our experience does demonstrate that push-based calls can still be quite useful. During the instrumented period from April 2020 to September 2021, 3.85 million push-based calls were answered by over 300,000 caller-ids. Out of these caller-ids, more than 120,000 were obtained exclusively through paper forms. Crucial information was pushed to the users about maternal and child health and nutrition, COVID-19 best practices and vaccination related awareness, AES (Acute Encephalitis Syndrome), and livelihood and awareness programmes conducted by the SHG network. An evaluation of the impact of this messaging is ongoing. Prior work conducted during 2016-2018 in the Nalanda pilot revealed significant gains in both awareness and practice, among both women who were reached through offline trainings as well as a distinct group of users who were engaged predominantly through push-based messaging \cite{jeevika}.

\begin{table*}[t]
  \centering
\begin{tabular}{|l|l|l|l|l|l|l|l|l|}
\hline
&\multicolumn{4}{c|}{\textbf{Nalanda}} & \multicolumn{4}{c|}{\textbf{Muzaffarpur}}\\
\hline
&\multicolumn{2}{c|}{\vtop{\hbox{\strut\textbf{Call-in}}\hbox{\strut\textbf{users}}}}&
\multicolumn{2}{c|}{\vtop{\hbox{\strut\textbf{OCR}}\hbox{\strut\textbf{users}}}}& 
\multicolumn{2}{c|}{\vtop{\hbox{\strut\textbf{Call-in}}\hbox{\strut\textbf{users}}}}&
\multicolumn{2}{c|}{\vtop{\hbox{\strut\textbf{OCR}}\hbox{\strut\textbf{users}}}}\\
\hline
\textbf{Number of unique callerids} & 
\multicolumn{2}{c|}{130390}& 
\multicolumn{2}{c|}{57423}& 
\multicolumn{2}{c|}{47330}&
\multicolumn{2}{c|}{65099}\\
\hline
\vtop{\hbox{\strut\textbf{Number of push-calls during}}\hbox{\strut\textbf{Apr 2020 to Sep 2021}}} & 
\multicolumn{2}{c|}{525564}&
\multicolumn{2}{c|}{1253577}&
\multicolumn{2}{c|}{422096}&
\multicolumn{2}{c|}{1653660}\\
\hline
\vtop{\hbox{\strut\textbf{Number of pull-calls during}} \hbox{\strut\textbf{Apr 2020 to Sep 2021}}} & 
\multicolumn{2}{c|}{68991}&
\multicolumn{2}{c|}{63144}& 
\multicolumn{2}{c|}{28479}& 
\multicolumn{2}{c|}{17109}\\
\hline
 & &   &            &             &                &         &           & \\
 \hline
& \textbf{Mean}&
\vtop{\hbox{\strut\textbf{Std}} \hbox{\strut\textbf{dev}}}&
\textbf{Mean}&
\vtop{\hbox{\strut\textbf{Std}} \hbox{\strut\textbf{dev}}}&
\textbf{Mean}&
\vtop{\hbox{\strut\textbf{Std}} \hbox{\strut\textbf{dev}}}&
\textbf{Mean}&
\vtop{\hbox{\strut\textbf{Std}} \hbox{\strut\textbf{dev}}} \\
\hline
\% \textbf{OBDs answered}& 18.5\%        & 4.9\%   & 34.3\%    & 4.1\%    & 14.4\%        & 3.8\%   & 28.3\%    & 10.3\% \\
\hline
\textbf{Duration of push-calls}& 2.2           & 1.8     & 2.8       & 2.1      & 2.4           & 1.9     & 2.5       & 2.0 \\
\hline
\vtop{\hbox{\strut\textbf{Percentage of OCR users}}\hbox{\strut\textbf{who did pull-calls}}}       &               &         & 4.2\%     & 1.8\%   &         &  & 1.6\%     & 0.9\%   \\
\hline
\textbf{Duration of pull-calls}                           & 8.4           & 5.2     & 10.2      & 6.6      & 6.7           & 3.8     & 6.5       & 3.7 \\
\hline

\end{tabular}
\\
\caption{\label{tab:table-3}
Project implementation efficacy for different types of users}

\end{table*}

\section{Conclusions}

In this study, we demonstrated the role that paper-based data collection can play in extending the outreach of voice-based communication for awareness of health practices among rural women in India. We were able to collect phone numbers of a large set of Self Help Group members in two districts of the state of Bihar in India, and successfully pushed numerous calls to them with crucial awareness information, especially during the COVID-19 pandemic. We contribute a unique dataset of handwritten digits that is more representative of real-world characteristics than the MNIST and EMNIST datasets, and a deep learning model and methodology for the digitization of this data. All data, models, and code for this effort have been released in the open-source. These resources can be applied to paper-based data collection in wider contexts as well, such as when field workers are required to provide data but may not have access or the capability to use digital devices directly for data entry. We show that hand-filled paper forms which are scanned and digitized automatically is a viable alternate low-cost pathway. We also plan to extend this work in the future so that point-of-service data entry can be done by the field workers by taking a photograph of the paper forms as and when they are filled, rather than physically transport them to a facility where the forms can be scanned in bulk. 

\begin{acks}
We would like to thank the Microsoft AI for Humanitarian Action and the IIT Delhi High Performance Computing (HPC) teams for providing us with the necessary infrastructure to conduct this research.  We also thank MEITY (Government of India) for providing a grant through the NLTM Bhashini project.
\end{acks}
\bibliographystyle{ACM-Reference-Format}
\bibliography{main}

\appendix

\end{document}